\pdfoutput=1

\documentclass[11pt]{article}

\usepackage{EMNLP2023}
\usepackage{lipsum} 
\usepackage{tcolorbox}
\usepackage{threeparttable}
\usepackage{times}
\usepackage{latexsym}
\usepackage[T1]{fontenc}
\usepackage[utf8]{inputenc}
\usepackage{microtype}
\usepackage{inconsolata}
\usepackage{CJKutf8}
\usepackage{graphicx}
\usepackage{array}
\usepackage{arydshln}
\usepackage{multirow}

\usepackage{bm}
\usepackage{amssymb}
\usepackage{amsmath}
\usepackage{makecell}
\usepackage{wrapfig}
\usepackage{lipsum}  
\usepackage{etoolbox}
\usepackage{comment}
\usepackage{enumitem}

\usepackage{booktabs}
  
\usepackage{microtype}
\usepackage{csquotes}
\usepackage{bm}
\usepackage{amsfonts,mathtools}
\usepackage{soul}
\usepackage{adjustbox}
\usepackage[capitalize]{cleveref}
\usepackage{xcolor}
\usepackage{hyperref}
\usepackage{appendix}
\usepackage{svg}
\AtBeginEnvironment{appendices}{\crefalias{section}{appendix}}
\BeforeBeginEnvironment{appendices}{\clearpage}

\Crefname{appsec}{appendix}{appendices}

\usepackage{caption}
\usepackage{subcaption}
\usepackage{enumerate}
\usepackage{scalerel}
\usepackage[ruled,vlined]{algorithm2e}
\usepackage{linguex}
\usepackage{relsize}
\usepackage{array}
\usepackage{booktabs}
\usepackage{array}
\usepackage{makecell}
\usepackage{relsize}
\usepackage{multirow}

\usepackage[font=small,labelfont=bf]{caption}

\usepackage{cancel}
\usepackage{inconsolata}

\usepackage[utf8]{inputenc}
\definecolor{darkorange}{rgb}{1, 0.549, 0}
\BeforeBeginEnvironment{appendices}{\clearpage}

\usepackage{todonotes}

\title{Stephanie: Step-by-Step Dialogues for Mimicking Human Interactions \\in Social Conversations}

\author{Hao Yang$^1$\thanks{~~Work done during an internship at FaceMind Corporation}~, 
  Hongyuan Lu$^2$\thanks{~~Corresponding author and co-first author},
  Xinhua Zeng$^1$,
  Yang Liu$^{1,3}$,
  Xiang Zhang$^4$,\\
  \textbf{Haoran Yang$^2$,
  Yumeng Zhang$^5$,
  Shan Huang$^4$,
  Yiran Wei$^4$,
  Wai Lam$^2$}\\
  $^1$Fudan University, $^2$The Chinese University of Hong Kong,  $^3$University of Toronto, \\ $^4$FaceMind Corporation, $^5$Tsinghua University \\
  \texttt{yanghao21@m.fudan.edu.cn, hongyuanlu@outlook.com}
}

\begin{document}
\maketitle 
\begin{abstract}
In the rapidly evolving field of natural language processing, dialogue systems primarily employ a single-step dialogue paradigm. Although this paradigm is efficient, it lacks the depth and fluidity of human interactions and does not appear natural. We introduce a novel \textbf{Step}-by-Step Dialogue Paradigm (Stephanie), designed to mimic the ongoing dynamic nature of human conversations. By employing a dual learning strategy and a further-split post-editing method, we generated and utilized a high-quality step-by-step dialogue dataset to fine-tune existing large language models, enabling them to perform step-by-step dialogues. We thoroughly present Stephanie. Tailored automatic and human evaluations are conducted to assess its effectiveness compared to the traditional single-step dialogue paradigm. We will release code,  Stephanie datasets, and Stephanie LLMs to facilitate the future of chatbot eras.

\end{abstract}

\section{Introduction}

In the field of natural language processing, the research and development of dialogue systems continue to advance. Some systems aim to mimic human communication in daily life\footnote{https://nijigen.com.cn}\textsuperscript{,}\footnote{https://chatgpt.com/?model=gpt-4o}, providing a more natural user experience. However, these systems predominantly employ a Single-Step Dialogue Paradigm\citep{abbas4657124context, touvron2023llama, du2022glm, abdin2024phi, achiam2023gpt}, where the system provides a comprehensive, one-time response to each user input, quickly addressing user questions or needs. While this approach can deliver a wealth of information in a single response, it falls short in simulating the fluidity of real human conversations and attracting user engagement. In reality, daily human conversations are ongoing, dynamically evolving processes involving multiple topics \citep{9746464, butler2011temporal, 10102573, poria2019emotion}. However, the current Single-Step Dialogue Paradigm fails to fully capture this complexity and lacks in attracting users to engage in conversations.

\begin{figure}[t!]
\centering
\includegraphics[width=0.5\textwidth]{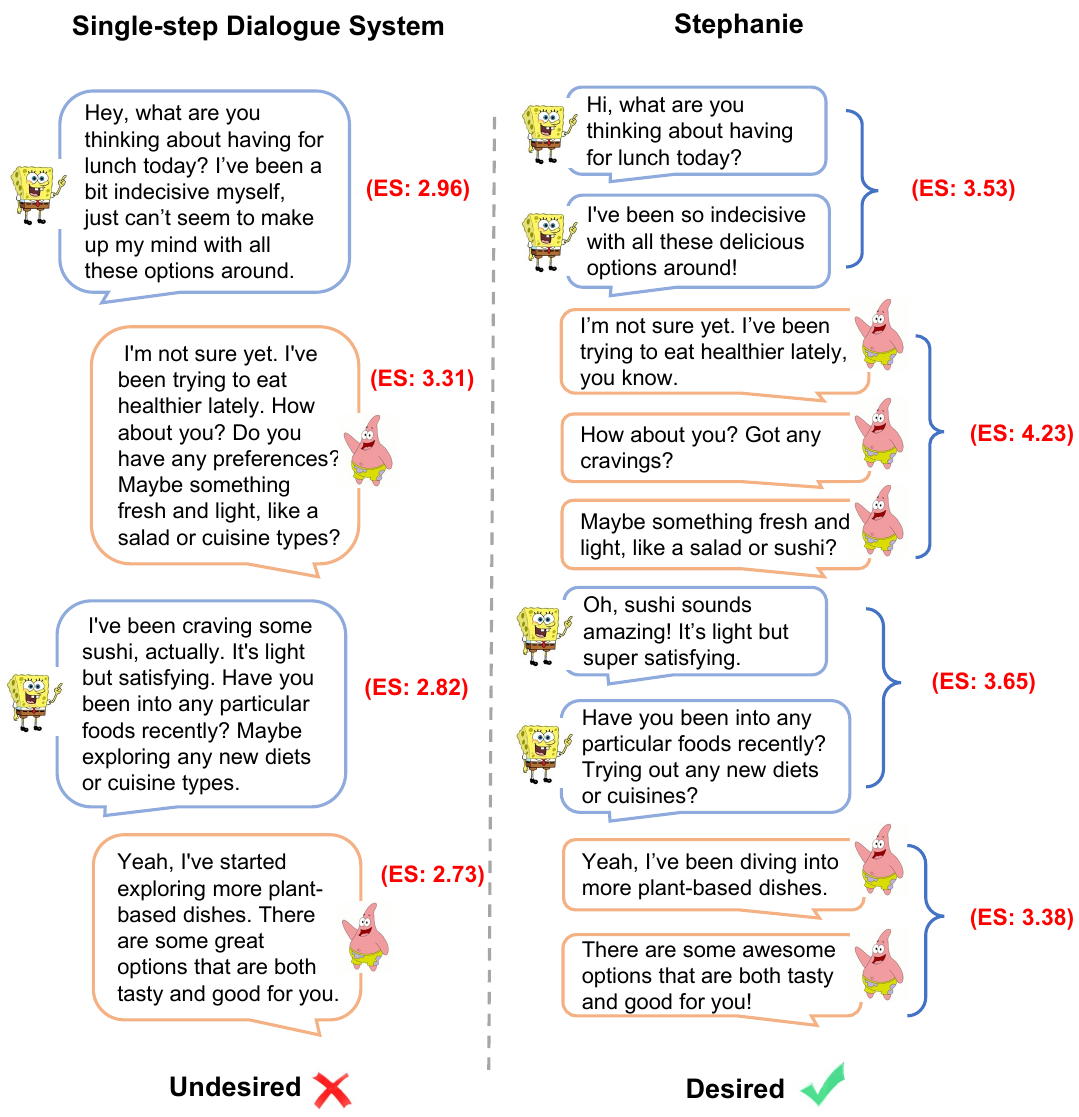}
\caption{A single-step dialogue system and Stephanie. Stephanie constructs a dialogue composed of multiple dispersed yet coherent responses. ES stands for the \textbf{engaging} score given by humans, which is introduced in section \ref{4.4}. }
\label{1-1}
\end{figure}

To better emulate the style of human social conversations, this paper introduces an innovative dialogue paradigm named step-by-step dialogue (Stephanie), where its human-rated engaging score surpasses that of single-step dialogue, as illustrated in Figure \ref{1-1}. Unlike single-step dialogue, Stephanie mimics casual chats in instant messaging applications, creating a more natural and continuous dialogue flow with the power of in-context learning \citep{min-etal-2022-metaicl, chen2023places, wang-etal-2024-learning}. Under this paradigm, the dialogue system does not just provide a one-time response to each input but constructs a conversation composed of multiple dispersed yet coherent responses. This design allows the system to gradually develop the conversation, with each response focusing on different aspects of the dialogue, making the conversation more detailed, rich, and engaging. We found that it provides a better conversation experience, evoking greater user engagement. For example, in the step-by-step chat mode, the system can address various aspects of a user’s expression step by step, first by supporting users through empathetic language and understanding, and then by asking questions or expanding the topic, gradually building deeper and more continuous communication.

If the one-time response of a single-step dialogue system is simply divided into multiple responses by punctuation, the overall logic and integrity of the one-time response itself will result in an unnatural and stiff step-by-step dialogue, which does not resemble a real social interaction with people. In order to fully consider the semantic similarities and differences between sentences, as well as naturalness and anthropomorphism when generating step-by-step dialogue and implementing a dialogue system with a step-by-step chat function, we introduced a comprehensive prompting framework that employs a dual learning strategy and a Further-Split post-editing method to generate and optimize step-by-step dialogue datasets. We then used this dataset with a specific fine-tuning strategy to be compatible with existing large models, thereby establishing a step-by-step dialogue system. The step-by-step dialogue paradigm demonstrates significant academic and practical value in enhancing the naturalness and engaging nature of chat systems. By simulating real social interactions, this research not only advances the technology of dialogue systems but also provides new insights and approaches for achieving more natural and human-like communication between machines and humans.

The main contributions of this paper include:
\begin{itemize}
\item We innovatively propose a step-by-step dialogue paradigm that utilizes a series of dispersed yet coherent responses to more closely mirror the style of real human communication interactions, thereby enhancing the engaging nature and human-likeness of the dialogue.

\item We introduced a bidirectional learning strategy and a Further-Split post-editing method to generate and optimize step-by-step dialogue datasets, and then we fine-tuned existing large models to develop a step-by-step dialogue system. To facilitate future research, we will release code, Stephanie datasets, and Stephanie LLMs in the near future. 

\item Finally, we comprehensively compared single-step dialogues with progressive dialogues through both human and automated evaluations, demonstrating the significant advantages of step-by-step dialogue systems over traditional single-step dialogue systems.

\end{itemize}

\section{Related Work}

\textbf{Large Language Models for Dialogue Systems} In dialogue systems, previous dialogue systems have been traditionally finetuned on publicly available dialogue datasets \citep{zhang2019dialogpt, adiwardana2020towards, roller2020recipes, thoppilan2022lamda}. Motivated by ChatGPT's success, developers are now conducting supervised finetuning on open-source large language models like LLaMA \citep{touvron2023llama} to develop dialogue systems. This process involves finetuning with constructed instruction-following examples\citep{taori2023stanford} and using dialogue data distilled from ChatGPT \citep{ulmer2024bootstrapping, chiang2023vicuna}. Furthermore, some studies have been prompting dialogue systems built on large pre-trained models to induce the knowledge embedded in these language models. Areas of focus include task-oriented dialogues \citep{labruna2023unraveling, swamy2023contextual, mi2022cins}, knowledge-supported dialogues \citep{semnani2023wikichat, rogers2023proceedings, hongru2023large}, and open-domain dialogues \citep{chen2023places, lee2023prompted, hongru2023cue}.

\noindent \textbf{Emotional Support in Dialogue Systems}
The role of emotions in building attractive dialogue systems has been thoroughly investigated \citep{zhou2017mojitalk, huber2018emotional, huang2020challenges}. Emotional chatting refers to dialogue systems expressing emotions such as happiness or sadness, while emotional support goes further, aiming to alleviate users' emotional distress in emotional chatting by proactively guiding the conversation and employing appropriate support techniques \citep{ratican2023six, chen2023llm, liu2023chatcounselor}.
An empathetic response is a key element in providing effective emotional support, focusing on understanding the user's emotions and making suitable replies, with the goal of creating more personalized and engaging responses\citep{liao2021dialogue, sun2021psyqa, majumder2020mime}. Enhancing the empathetic response capability of LLMs through context learning with semantic similarity, bi-directional co-generation, and integration with knowledge bases has been proposed \citep{qian2023harnessing}. Additionally, intermediate reasoning steps can be adopted, using language clues in the conversation to determine the user's emotional state, personality traits, and psychological characteristics, and then generating empathetic responses \citep{hongru2023cue}. Emotional support dialogue also requires the ability to devise dialogue strategies, formulate appropriate response strategies for various emotional problems of users, and achieve complex dialogue objectives such as exploration, comfort, and action \citep{rogers2023proceedings, peng2022control, tu2022misc}. A multi-agent framework can coordinate multiple specialized agents, each responsible for a specific aspect of complex dialogue objectives in emotional support, such as exploration, comfort, and action, making complex dialogue objectives more approachable and stimulating greater intelligence through collaboration \citep{cheng2023cooper}. Another dialogue strategy involves breaking down the ultimate goal into a sequence of sub-goals, selecting actions for sub-goals, and filtering valuable sub-goals to efficiently achieve the ultimate goal \citep{chua2024towards}. Currently, emotional support based on LLMs faces the challenge of data scarcity. One approach is to use dialogues as generative seeds and exploit the contextual learning potential of ChatGPT to recursively generate scalable emotional support dialogue datasets \citep{zheng2023building}.

In the current field of natural language processing, most dialogue systems based on large language models primarily adopt a Single-Step Chat Paradigm\citep{wu2023brief, touvron2023llama, mai2023llm, yamazaki2023building}. Within this paradigm, the system responds to each user input with a comprehensive and complete one-time reply to promote interaction. Such interactions provide information-dense responses to handle complex inquiries, focusing on the informational density and completeness of each response, which is suitable for directly resolving specific questions or providing detailed information in a single interaction. However, this paradigm exhibits certain limitations in emulating the natural fluidity and emotional expression found in human daily dialogues. While it can identify and respond to users' emotional inclinations, the interaction pattern often sticks to a question-and-single-answer format, lacking the emotional continuity and interaction depth present in real conversations. 

\section{Methodology}

\begin{figure*}[ht!]
\centering
\includegraphics[width=\textwidth]{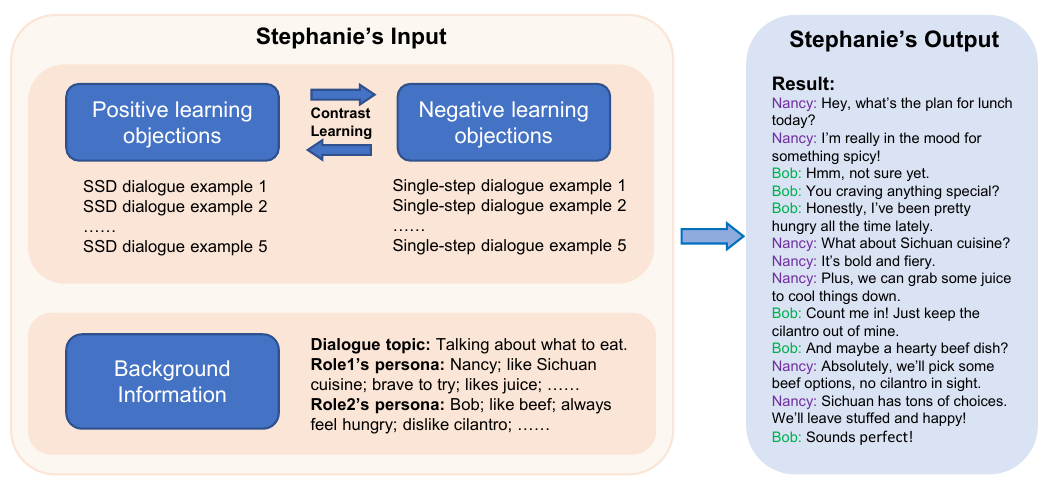}
\caption{In the process of step-by-step dialogue generation, we adopted a dual learning strategy to enhance the model's ability to generate natural dialogues through the Step-by-Step Dialogue Prompt Framework. This strategy combines positive and negative learning objectives. The positive objective includes high-quality step-by-step dialogue examples selected from real social interactions, while the negative objective comprises designed high-quality single-step dialogue examples. Through contrastive learning, this approach helps the model distinguish between step-by-step dialogues and single-step dialogues, thus generating more natural and emotionally rich step-by-step dialogues.}
\label{3-1}
\end{figure*}

In this section, we will delve into the process of generating and optimizing step-by-step dialogues, and based on this, create a high-quality step-by-step dialogue dataset. We further fine-tuned and built a dialogue system capable of step-by-step interactions to simulate the step-by-step dialogue paradigms found in real human social exchanges.

\subsection{Dual Learning Strategy for Step-by-Step Dialogue Generation}

To efficiently generate step-by-step dialogues that mimic real human social interactions, inspired by contrastive learning, we propose a dual learning strategy combining both positive and negative learning objectives within a comprehensive prompt framework named step-by-step dialogue prompt framework. As illustrated in Figure \ref{3-1}, the framework consists of three elements: background information $D$, positive learning objectives $P$, and negative learning objectives $N$, aiming to enhance the model’s ability to generate dialogues that are both rich and natural. the comprehensive prompt framework can be formulated as:
\begin{equation}
p(r\mid D, P, N)
\end{equation}

where $r$ is the response output of the model, and the design of the three elements is as follows:

\begin{itemize}
\item \textbf{Background Information:} We use an LLM to summarize and generate the themes \( T \) of each dialogue segment from the persona-chat dataset and the characteristics \( C \) of the dialogue participants, to form the background information \( D = \{T, C\} \). This information guides the model's generation, covering common topics such as family, work, and leisure activities, while considering the diverse personalities of the dialogue participants—for example, one might be described as optimistic and active, while another might be portrayed as having recently faced setbacks but remaining diligent and academically inclined.
\item \textbf{Positive Learning Objectives:} To help the model understand the step-by-step dialogue paradigm, we created five high-quality step-by-step dialogue examples as the positive objectives $P$. These examples simulate everyday social exchanges between two individuals and serve as a basis for few-shot learning, training the model to generate coherent and emotionally rich step-by-step dialogues in different background contexts.
\item \textbf{Negative Learning Objectives:} Simultaneously, we designed five high-quality single-step dialogue examples as the negative objectives $N$, which share the same theme as the five step-by-step dialogue examples for the positive objectives. Through contrastive learning, these examples enable the model to discern the differences between single-step and step-by-step dialogues. This negative learning approach helps the model better understand the step-by-step dialogue paradigm by pushing away dissimilar single-step dialogue examples.
\end{itemize}

This dual learning strategy is a robust prompting framework. Through this structured approach, the model considers both positive and negative learning objectives during dialogue generation, enhancing its ability to understand and generate step-by-step dialogues while ensuring that the generated dialogues align with the themes and character traits in the background information.

\subsection{Optimizing Step-by-Step Dialogues Using the Further-Split Post-Editing Method}

Although the method described in Section 3.1 enabled the model to make some progress in generating step-by-step dialogues, our evaluation showed that some generated dialogues still exhibited characteristics of single-step dialogues, such as dense, one-time responses. To address this issue and further enhance the coherence and naturalness of emotional expression in dialogues, we designed a post-editing optimization method called "further-split."

In this process, we selected five initial step-by-step dialogues generated by the model for detailed analysis and manual restructuring. We further split these dialogues according to the natural flow and emotional progression of actual conversations, reorganizing and optimizing the content. The optimized step-by-step dialogue examples were paired with the original examples, serving as rewritten examples to guide the model in learning how to further split and rewrite dialogues, thereby generating more natural and human-like step-by-step dialogues to closely mimic real social interactions.

\subsection{Dataset Generation and Finetuning Strategy for Stephanie}

Based on the aforementioned comprehensive prompt framework and the further-split post-editing method, we generated a high-quality step-by-step dialogue dataset. To effectively utilize this dataset for finetuning existing large language models, we designed a specific finetuning strategy.

During the finetuning process, we introduced delimiters to format the dataset, providing structured input and output for the model, where the content between each pair of delimiters represents a single exchange between the dialogue participants. We then used this newly formatted step-by-step dialogue dataset to finetune the model. After finetuning, the model’s output also adopted the same delimiter-separated step-by-step dialogue format. Then, we found Stephanie brings better user engagement, converting the delimiter format of the input and output into a format similar to message bubbles in social software, allowing users to interact with the large language model using the step-by-step dialogue paradigm.

This plug-and-play finetuning strategy enables our step-by-step dialogue dataset to be compatible with various existing language models, thereby constructing a dialogue system capable of step-by-step interactions to provide a coherent and emotionally rich dialogue experience in practical applications.
\section{Experiment Setup}

\subsection{Dataset}
Our incremental dialogue dataset originates from the PERSONACHAT dataset. It is a renowned multi-turn dialogue dataset grounded in character personas, with each dialogue instance typically comprising around 8 turns, where each self and partner character is described by roughly 4 traits.

From the training set of PERSONA-CHAT, we curated 5,457 high-quality dialogues. Initially, we employed the Llama3-70b model to summarize the theme of each dialogue, with summaries averaging between 50 to 100 words. Subsequently, we adopted the Stephanie dialogue generation approach described herein, incorporating these dialogue themes and approximately 4 traits of both characters as background information. We use the Llama3-70b model to generate an incremental dialogue dataset consisting of 5,457 dialogues, where each character involved exhibits roughly 4 traits. 

\subsection{Prompt}
We describe the prompt generated for step-by-step dialogue with Stephanie as follows:

\begin{tcolorbox}[colframe=gray, colback=white, boxrule=1pt, sharp corners]
\textit{<five examples of single-step dialogues>.\\
<five examples of step-by-step dialogues>.\\
In single-step dialogues, each role sends only one message per turn. In contrast, step-by-step dialogues allow multiple messages to be sent consecutively before the other role replies, simulating the style of human daily chit-chat. Please generate a step-by-step dialogue and a single-step dialogue based on the background information:\\
<background information>.}
\end{tcolorbox}

We can also describe the prompt optimized for generating step-by-step dialogue using the Further-Split method as follows:

\begin{tcolorbox}[colframe=gray, colback=white, boxrule=1pt, sharp corners]
\textit{<five examples of single-step dialogues and corresponding Stephanie>.\\
Please assess whether each message reply in the following step-by-step dialogue can be further rewritten into multiple replies to make the conversation more natural, interesting, engaging, and closer to human interaction. Then, provide a new version of the step-by-step dialogue:\\
<the single-step dialogue to be further-split into Stephanie>.}

\end{tcolorbox}

\subsection{Baselines and Comparison Models}
In evaluating the performance of our model, we consider several leading models in the field of language processing. These models are used as benchmarks due to their significant capabilities in various tasks within natural language processing. Each model is briefly described as follows:

\begin{itemize}
\item \textbf{GPT-4}: Developed by OpenAI, GPT-4 represents the latest advancement in the Generative Pre-trained Transformer series. Renowned for its vast knowledge base and flexibility across multiple tasks, GPT-4 is a critical benchmark for assessing advanced language understanding and generation capabilities.

\item \textbf{Llama3-70b}: Also from Meta's Llama series, the Llama3-70b model, with its 70 billion parameters, is aimed at deep contextual understanding and complex reasoning tasks. It serves as a high-end model for performance comparison.

\item \textbf{Llama3-8b}: A model from Meta's Llama family, Llama3-8b is designed to provide a balance between performance and efficiency with its 8 billion parameters. It is optimized for rapid response and lower resource usage, making it suitable for real-time applications.

\item \textbf{Phi3-3.8b}: Phi3-3.8b from Microsoft's Phi-3 series of small language models excels in performance while being highly efficient in terms of computational resource usage. These models are designed for flexible deployment across cloud, on-device, and edge computing scenarios, ensuring effectiveness even with limited connectivity. Phi3-3.8b uses high-quality, curated training data to achieve results comparable to larger models.
\end{itemize}

\subsection{Evaluation Metrics} \label{4.4}

To comprehensively assess the performance of our step-by-step dialogue, we have utilized a series of evaluation metrics aimed at thoroughly measuring various aspects such as the diversity, naturalness, and effectiveness of the dialogues, among others. These metrics include: Dialogue Experience Metrics (suitable for both automated and human evaluations), Lexical Diversity Metrics (Distinct-N), and statistical features of the dialogue data, such as the average number of words per message and the Average Consecutive Message Counts (ACMC).

\begin{itemize}

\item \textbf{Dialogue Experience Metrics}:
\textbf{Interesting}: The degree of interest in the dialogue. If the dialogue carries a negative sentiment, the score is 0.
\textbf{Informative}: The amount of information contained in the dialogue.
\textbf{Natural}: Whether the dialogue is natural and human-like.
\textbf{Engaging}: Whether the dialogue is engaging, meaning if what is said by both roles makes them want to continue the dialogue.
\textbf{On-topic}: Whether the dialogue stays on the topic described in the dialogue topic.
\textbf{On-persona}: Whether the dialogue matches the personas of role1 and role2.

\item \textbf{Distinct-N}:
To quantify the lexical diversity of the dialogues, we utilize the Distinct-N metric. This metric calculates the diversity of n-grams in the generated responses across all possible values of \( N \), showing the system's capability to produce varied and engaging content. The Distinct-N is defined as:
\[
\text{Distinct-N} = \frac{\text{Total unique n-grams}}{\text{Total n-grams}}
\]

\item \textbf{Words/Message}:
Calculates the average number of words per message, providing insight into the verbosity or conciseness of the dialogues. This helps in determining the efficiency and clarity of the communication. The formula for Words/Messages is defined as:

\[
\text{Words/Message} = \frac{\sum_{i=1}^{n} w_i}{n}
\]

where \( w_i \) is the number of words in the \( i \)-th message, and \( n \) represents the total number of messages.

\item \textbf{ACMC (Average Consecutive Message Counts)}:
This metric measures the average number of consecutive messages sent by one participant before receiving a response. It is calculated as:
\[
\text{ACMC} = \frac{\sum_{i=1}^{n} c_i}{m}
\]
where \( c_i \) is the number of consecutive messages in the \( i \)-th turn without interruption by the other participant, \( n \) is the total number of such turns, and \( m \) is the total number of messages sent by the participant.
\end{itemize}

\section{Results}

\begin{table*}[h!t]
\begin{threeparttable}
\begin{tabular*}{\textwidth}{@{\extracolsep\fill}lcccccccc}
\toprule
& \multicolumn{4}{c}{GPT4} & \multicolumn{4}{c}{Llama3-70b}\\
\cmidrule(lr){2-5}\cmidrule(lr){6-9}
Metrics & \textbf{\( \alpha \)} & \textbf{\( \beta \)} & \textbf{\( \gamma \)} & Stephanie & \textbf{\( \alpha \)} & \textbf{\( \beta \)} & \textbf{\( \gamma \)} & Stephanie \\
\midrule
Interesting   & 82.00 & 80.00 & 84.66 & \textbf{88.35} & 80.20 & 75.25 & 88.74 & \textbf{91.63} \\
Informative     & 83.20 & 80.35 & 85.03 & \textbf{88.19}  & 79.24 & 72.93 & 86.67 & \textbf{88.29} \\
Natural & 87.25 & 88.19 & 91.89 & \textbf{94.72} & 86.82 & 84.36 & 95.00 & \textbf{97.61}\\
Engaging   & 85.74 & 84.84 & 89.42 & \textbf{92.64} & 83.64 & 78.74 & 93.25 & \textbf{95.86} \\
On-topic     & -- & 91.54 & 93.53 & \textbf{96.35} & -- & 87.78 & 95.88 & \textbf{97.10} \\
On-persona    & -- & 92.93 & 94.25 & \textbf{96.0} & -- & 90.05 & 96.53 & \textbf{98.10} \\
\bottomrule
\end{tabular*}
\caption{Automatic Evaluation on GPT4 and Llama3-70b. The values represent the percentage scores for each metric, used to evaluate the performance of different dialogues (\textbf{\( \alpha \)}, \textbf{\( \beta \)}, \textbf{\( \gamma \)}, Stephanie) generated by GPT-4 and Llama3-70b. These scores indicate how interesting, engaging, informative, and natural each dialogue is, as well as its adherence to the given topic and persona. Bold values indicate the highest scores among comparable models, highlighting exceptional performance in specific metrics.}\label{tab1}
\end{threeparttable}
\end{table*}

\begin{table}[h!t]
\begin{minipage}{0.5\textwidth}
\begin{threeparttable}
\begin{tabular*}{\textwidth}{@{\extracolsep\fill}lccc}
\toprule
Metrics & \textbf{\( \alpha \)} & \textbf{\( \beta \)} & Stephanie \\
\midrule
Interesting & 79.73 & 72.14 & \textbf{83.53} \\
Informative & 75.48 & 74.56 & \textbf{79.37} \\
Natural & 79.79 & 75.87 & \textbf{87.41} \\
Engaging & 83.55 & 78.38 & \textbf{86.41} \\
On-topic & -- & 78.87 & \textbf{82.57} \\
On-persona & -- & 77.24 & \textbf{80.04} \\
\bottomrule
\end{tabular*}
\caption{Automatic evalution on phi3-3.8b. The table shows the percentage scores of the performance of three dialogues (\textbf{\( \alpha \)}, \textbf{\( \beta \)}, \textbf{\( \gamma \)}) across multiple metrics. Bold values represent the best performance in each metric for the phi3-3.8b evaluation.}\label{tab2}
\end{threeparttable}
\end{minipage}
\end{table}

\begin{figure}[t]
\centering
\includegraphics[width=0.5\textwidth]{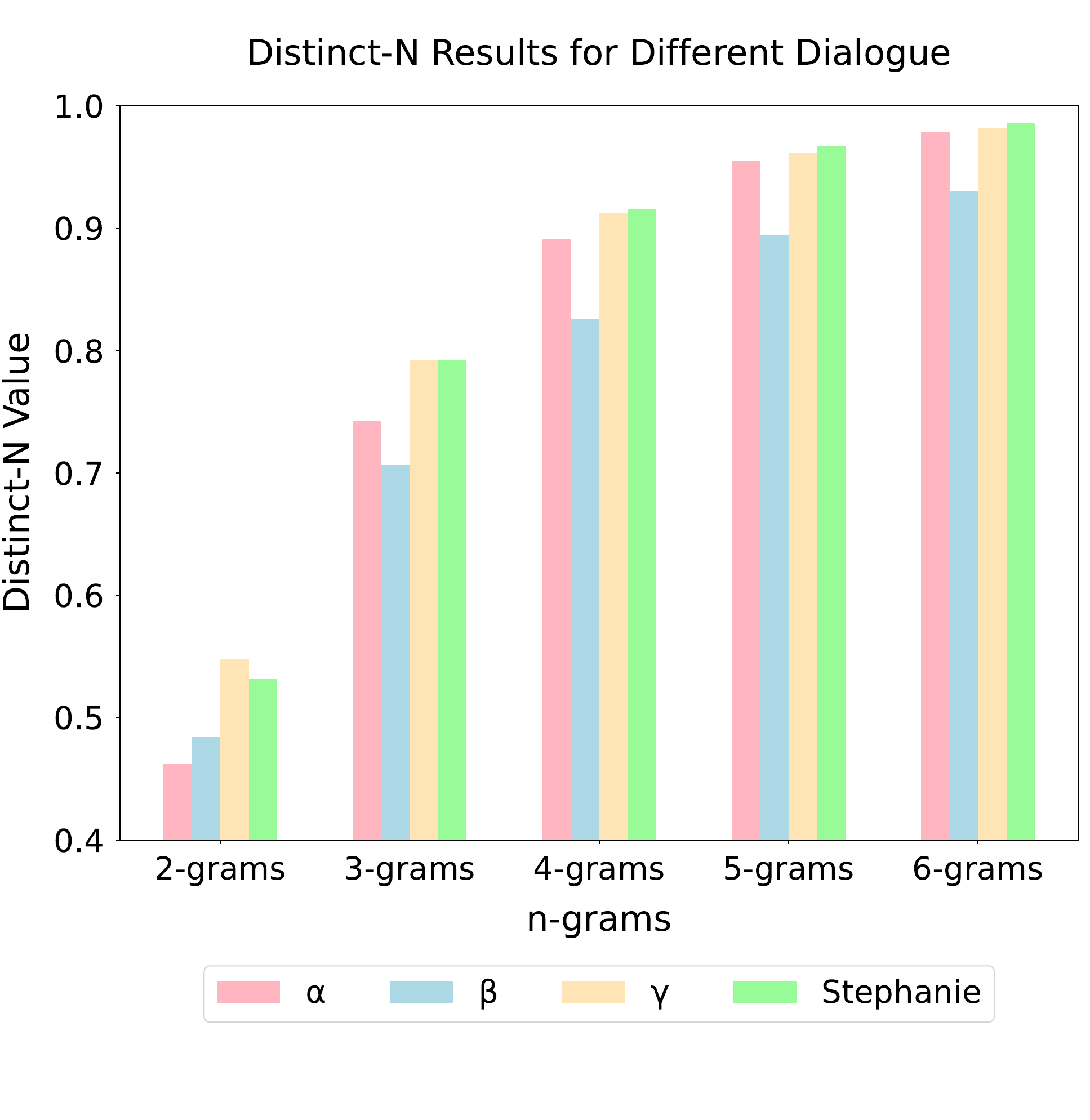}
\caption{Distinct-N Results for Different Dialogue. This graph displays the lexical diversity of dialogues generated by various models, measured by the Distinct-N metric for n-grams from N=2 to N=6. Each colour represents a different dialogue model (\textbf{\( \alpha \)}, \textbf{\( \beta \)}, \textbf{\( \gamma \)}, Stephanie), highlighting variations in linguistic complexity and diversity.}
\label{5-1}
\end{figure}

\subsection{Evaluation of Conversation Quality}

We selected 100 conversation data from the persona-chat dataset as the Original Single-Step Dialogue \textbf{\( \alpha \)}. First, we used GPT-4 to summarize the themes of these 100 dialogues. Then, along with the personas of the dialogue participants, we used this background information to write prompts using the Step-by-Step Dialogue Prompt Framework proposed in this paper. These prompts were fed into GPT-4 to generate the Generated Single-Step Dialogue  \textbf{\( \beta \)}. We also applied a further-split method to optimize the \textbf{\( \beta \)}, resulting in the Further-Split Step-by-Step Dialogue \textbf{\( \gamma \)}. Additionally, we conducted corresponding experiments with the Llama3-70b and phi3-3.8b models, generating their respective \textbf{\( \alpha \)}, \textbf{\( \beta \)}, \textbf{\( \gamma \)}, and Stephanie.

Subsequently, we conducted automatic machine assessments of the three models on six metrics: Interesting, Informative, Natural, Engaging, On-topic, and On-persona. with Claude-3-sonnet as the assessment expert providing scores from 0 to 100, as shown in tables \ref{tab1} and \ref{tab2}. The \textbf{\( \beta \)}s generated by the large models were generally weaker than the original dialogues on most metrics, with the exception of the 'Natural' metric for GPT-4, where \textbf{\( \beta \)} performed better than \textbf{\( \alpha \)}. This indicates that single-step dialogues generated by large models are inferior to original human dialogues. The \textbf{\( \gamma \)} was significantly superior to \textbf{\( \alpha \)} on all six metrics, demonstrating the superiority of the step-by-step dialogue paradigm. Stephanie showed further improvement over the \textbf{\( \beta \)}, highlighting the effectiveness of the further-split method. Additionally, we conducted a human evaluation of GPT-4, inviting three advanced graduate students majoring in English to score on a 0-5 scale. The results were positively consistent with the prior results.

\begin{table}[t]
\begin{minipage}{0.5\textwidth}
\begin{threeparttable}
\begin{tabular*}{\textwidth}{@{\extracolsep\fill}lcccccc}
\toprule
Metrics & \textbf{\( \alpha \)} & \textbf{\( \beta \)} & \textbf{\( \gamma \)} & Stephanie \\
\midrule
Interesting & 2.93 & 2.85 & 3.53 & \textbf{3.68} \\
Informative & 3.71 & 3.13 & 3.78 & \textbf{3.91} \\
Natural & 2.97 & 2.89 & 3.65 & \textbf{3.97} \\
Engaging & 3.13  & 2.96 & 3.72 & \textbf{4.06} \\
On-topic & -- & 3.30 & 3.79 & \textbf{3.99} \\
On-persona & -- & 3.17 & 3.73 & \textbf{3.89} \\
\bottomrule
\end{tabular*}
\caption{Human evalution on GPT4. The table presents human evaluation scores for different dialogue models (\textbf{\( \alpha \)}, \textbf{\( \beta \)}, \textbf{\( \gamma \)}, Stephanie) generated by GPT-4. Scores range from 1 to 5, with higher scores indicating better performance. }\label{tab3}
\end{threeparttable}
\end{minipage}
\end{table}

We conducted further statistics on the \textbf{\( \beta \)}, \textbf{\( \gamma \)}, and Stephanie generated by GPT-4. Table 4 presents the statistics for \textbf{\( \alpha \)}, \textbf{\( \beta \)}, \textbf{\( \gamma \)}, and Stephanie, including the average number of words per response (Words/Messages) and the Average Number of Consecutive Message Counts (ACMC). The results show that the \textbf{\( \beta \)} is similar to \textbf{\( \alpha \)}, with \textbf{\( \beta \)}'s Words/message being slightly higher than \textbf{\( \alpha \)}'s. Compared to \textbf{\( \alpha \)} and \textbf{\( \beta \)}, \textbf{\( \gamma \)} has fewer Words/message and a higher ACMC, indicating that step-by-step dialogues tend to be shorter and contain more messages. Notably, Stephanie, in comparison to \textbf{\( \gamma \)}, further effectively reduces Words/Messages and significantly increases ACMC, demonstrating the effectiveness of the further-split method. Table \ref{tab5} displays the proportion of consecutive message counts, where it is also evident that \textbf{\( \gamma \)}, compared to \textbf{\( \alpha \)} and \textbf{\( \beta \)}, has more consecutive replies. Furthermore, Figure \ref{5-1} illustrates that Stephanie effectively shifts the distribution of the number of consecutive messages to the right relative to \textbf{\( \gamma \)}.

\begin{table}[t]
\begin{minipage}{0.5\textwidth}
\begin{threeparttable}
\begin{tabular*}{\textwidth}{@{\extracolsep\fill}lccccc}
\toprule
Metrics & \textbf{\( \alpha \)} & \textbf{\( \beta \)} & \textbf{\( \gamma \)} & Stephanie \\
\midrule
words/message & 11.77 & 13.67 & 8.12 & 5.87 \\
ACMC & 1.07 & 1.08 & 1.99 & 2.51 \\
\bottomrule
\end{tabular*}
\caption{Words/message and ANT on dialogues. The table compares the average words per message and the Average Number of Consecutive Message Counts (ACMC) across different dialogue (\textbf{\( \alpha \)}, \textbf{\( \beta \)}, \textbf{\( \gamma \)}, Stephanie). This helps in evaluating the verbosity and interaction depth of each dialogue.}\label{tab4}
\end{threeparttable}
\end{minipage}
\end{table}

\begin{table}[t]
\begin{minipage}{0.5\textwidth}
\begin{threeparttable}
\begin{tabular*}{\textwidth}{@{\extracolsep\fill}lcccccc}
\toprule
Dialogues & one & two & three & four & five \\
\midrule
\textbf{\( \alpha \)} & 92.65  & 7.35 & 0 & 0 & 0 \\
\textbf{\( \beta \)} &  91.26 &  8.74 & 0 & 0 & 0 \\
\textbf{\( \gamma \)} & 20.50  & 60.10 & 17.98 & 1.21 & 0.1 \\
Stephanie & 11.17  &  39.24 & 34.33 & 10.86 & 2.97\\
\bottomrule
\end{tabular*}
\caption{The proportion of consecutive message counts. The table shows the proportion of dialogues with a given number of consecutive messages (one, two, three, four, five) for different dialogue models (\textbf{\( \alpha \)}, \textbf{\( \beta \)}, \textbf{\( \gamma \)}, Stephanie). Higher counts indicate a greater tendency for step-by-step dialogues within an interaction.}\label{tab5}
\end{threeparttable}
\end{minipage}
\end{table}
\begin{table}[t]
\begin{minipage}{0.5\textwidth}
\begin{threeparttable}
\begin{tabular*}{\textwidth}{@{\extracolsep\fill}lcccccc}
\toprule
Metrics & Stephanie-Llama3-8b & Llama3-8b \\
\midrule
Interesting &  \textbf{3.67} & 3.01 \\
Informative &  \textbf{3.81} & 3.22 \\
Natural &  \textbf{4.13} & 3.57 \\
Engaging &  \textbf{3.89} & 3.31 \\
\bottomrule
\end{tabular*}
\caption{Human evalution on Stephanie-Llama3-8b dialogue system. The table presents human evaluation scores for the Stephanie-Llama3-8b and Llama3-8b dialogue systems across four metrics: Interesting, Informative, Natural, and Engaging. Scores range from 1 to 5, with higher scores indicating better performance. The fine-tuned Stephanie-Llama3-8b model outperforms the Llama3-8b model across all metrics.}\label{tab6}
\end{threeparttable}
\end{minipage}
\end{table}

In assessing lexical diversity among dialogue models, the "Distinct-N" table provides a comparative analysis using the Distinct-N metric for n-grams ranging from N=2 to N=6. As shown in fig \ref{5-1}, The Original Single-Step Dialogue \textbf{\( \alpha \)} maintains high diversity, which increases with the complexity of n-grams, reflecting typical human dialogue characteristics. However, the Generated Single-Step Dialogue \textbf{\( \beta \)} exhibits lower diversity scores, especially for higher n-grams, indicating limitations in linguistic variability. Notably, the Generated Step-by-Step Dialogue \textbf{\( \gamma \)} and Further-Split Step-by-Step Dialogue (Stephanie) show superior performance, with Stephanie achieving the highest diversity across most categories. The significant performance of Stephanie in larger n-grams highlights the effectiveness of the further-split method in producing dialogues that are diverse and closely mimic the complex linguistic structures of human communication. This demonstrates that our proposed generation methods and prompting framework can significantly enhance the quality and human-likeness of machine-generated text.

\subsection{Fine-Tuning with Step-by-Step Dialogue}

Following the demonstration of the effectiveness of our proposed paradigms, generation methods, and prompting frameworks, we aimed to provide a high-quality dataset for fine-tuning existing large models. To this end, we generated a high-quality step-by-step dialogue dataset consisting of 5,457 segments using the Llama3-70b model. Subsequently, we fine-tuned the Llama3-8b model with this dataset to create the Stephanie-Llama3-8b model. We engaged five testers to interact with both the Llama3-8b and Stephanie-Llama3-8b dialogue systems, assessing them across four metrics. The results show that the model fine-tuned with the step-by-step dialogue dataset exhibited superior step-by-step dialogue capabilities, outperforming the Llama3-8b model on all four metrics as presented in table \ref{tab6}.

\section{Conclusion}
The step-by-step dialogue paradigm introduced in this article enhances human-like interactions in simulated dialogue systems. By integrating a dual learning strategy and a further-split post-editing method, we have effectively generated dialogue data with Stephanie that is more interesting, natural, engaging, and emotionally nuanced. Our evaluations demonstrate that Stephanie's 
systems significantly outperform traditional single-step dialogue systems across various metrics. We plan to release our code, Stephanie dataset and Stephanie systems in the near future to facilitate chatbot eras.

\section*{Limitations}
We conducted manual testing with limited human resources, and we look forward to seeing the application effectiveness of this technology on more large-scale consumer products.

\section*{Ethics Statement}

We honour and support the ACL Code of Ethics.
The datasets used in this work are well-known and
widely used, and the dataset pre-processing does
not make use of any external textual resource. In
our view, there is no known ethical issue. End-to-end pre-trained generators are also used, which
are subjected to generating offensive context. However,
the above-mentioned issues are widely known to
commonly exist for these models. Any content
generated does not reflect the view of the authors.

\bibliography{ref}
\bibliographystyle{acl_natbib}

\appendix

\end{document}